\documentclass[journal]{IEEEtran}
\usepackage{cite}
\usepackage{amsmath,amssymb,amsfonts}
\usepackage{graphicx,adjustbox,booktabs,textcomp,placeins,flafter,float}
\usepackage[table]{xcolor}
\usepackage{url}
\usepackage[hidelinks]{hyperref}
\definecolor{Gray}{gray}{0.9}
\hypersetup{pdftitle={SPACE-CLIP: Spatial Perception via Adaptive CLIP Embeddings for Monocular Depth Estimation},pdfauthor={Taewan Cho, Taeryang Kim, Andrew Jaeyong Choi}}
\title{SPACE-CLIP: Spatial Perception via Adaptive CLIP Embeddings for Monocular Depth Estimation}
\author{Taewan Cho, Taeryang Kim, and Andrew Jaeyong Choi\textsuperscript{*}\\[0.6em]
{\small School of Computing, iRASC Lab., Gachon University, Republic of Korea}\\[-0.15em]
{\small\ttfamily\{taewan2002, ktr1110, andrewjchoi\}@gachon.ac.kr}%
\thanks{\textsuperscript{*}Corresponding author: Andrew Jaeyong Choi.}}
\begin{document}
\maketitle
\begin{abstract}
Robotic and autonomous systems need dense spatial cues, yet adding a dedicated depth estimator can duplicate visual processing already performed by a multimodal model. CLIP-based depth methods offer an alternative, but commonly rely on text-derived conditioning or backbone adaptation. We present SPACE-CLIP, a decoder-only framework for supervised monocular depth estimation with a frozen CLIP vision backbone and no text encoder at inference. A FiLM-conditioned semantic pathway combines global image context with multilevel patch features, while a structural pathway supplies separately processed spatial features to a hierarchical fusion decoder. Indoor and outdoor evaluations demonstrate depth reconstruction with this architecture, and controlled component comparisons support the contribution of the structural pathway. Layer-selection experiments and frequency interventions further characterize the structural pathway's contribution to depth reconstruction. A shared-backbone microbenchmark further illustrates the reduction in duplicated computation. SPACE-CLIP provides a modular approach to adding dense depth prediction to compatible visual perception stacks. Code is available at \url{https://github.com/taewan2002/SPACE-CLIP}.
\end{abstract}
\begin{figure}[!t]
\centering
\includegraphics[width=0.96\columnwidth]{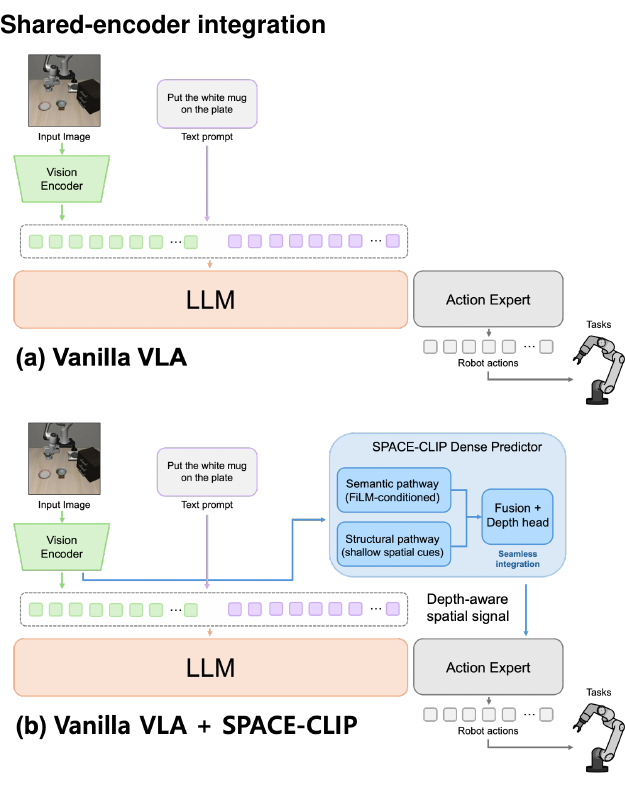}
\caption{Proposed shared-encoder integration of SPACE-CLIP. The depth module reuses a compatible vision encoder; downstream control is not evaluated.}
\label{fig:conceptual_paradigm}
\end{figure}
\section{Introduction}
\label{sec:introduction}
Large-scale vision-language models (VLMs), such as CLIP \cite{radford2021learning}, learn visual representations through image--text alignment and support recognition from natural-language descriptions. In robotic and autonomous systems, however, semantic recognition alone is not enough. Manipulation, navigation, and decision-making also depend on dense spatial cues that describe scene layout, object boundaries, and relative distance. Monocular depth estimation remains a practical route to such cues because it uses a single RGB stream, but it requires dense geometric reasoning and precise local structure.

Prediction quality is only part of the problem. In robotic and embodied systems, a depth module must also integrate with an existing perception-action stack. Monocular visual servoing and VLM-assisted grasping illustrate the connection between visual perception and action \cite{ras_ribeiro2021_visualservo,ras_kim2026_vlmgrasp}. Dedicated depth models, including DPT \cite{ranftl2021vision} and Depth Anything V2 \cite{yang2024depthanythingv2}, provide dense predictions through their own visual encoders and decoders. Running such a model alongside a vision-language-action (VLA) model can duplicate image encoding. Spatial context is also important in lightweight VLA design \cite{koo2025retovlareusingregistertokens}, motivating a systems question: can we add depth prediction by reusing the vision encoder already present in a compatible multimodal stack?

CLIP-based depth methods offer a starting point. DepthCLIP \cite{zhang2022can} and learned prompting \cite{Auty_2023_ICCV} connect image features to depth through language representations, while CaBins \cite{son2024cabins} adapts the image pathway and depth prompts. CLIP2Depth \cite{kim2024clip} conditions a decoder on a learned mirror embedding passed through a frozen text encoder. These approaches use text-derived representations, while methods that adapt the shared backbone also change features used by the host model. We target text-free inference with a frozen vision backbone (TFI-FB): all depth-specific learning is confined to a decoder, and depth prediction does not use a text encoder.

To meet these constraints, we propose SPACE-CLIP, a dual-pathway decoder that predicts depth from intermediate CLIP vision features. Figure~\ref{fig:conceptual_paradigm} illustrates the intended shared-encoder integration, and Fig.~\ref{fig:architecture} shows the depth model. The semantic pathway uses FiLM \cite{perez2018film} to condition multilevel features on global image context. The structural pathway processes a separate set of features without FiLM and supplies them to staged fusion blocks. This design combines context-conditioned decoding with spatial refinement while keeping the encoder fixed.

NYU Depth V2 \cite{Silberman:ECCV12} and KITTI \cite{Geiger2013IJRR} evaluations demonstrate indoor and outdoor depth reconstruction, and the decoder also operates with a frozen SigLIP backbone. Controlled NYU experiments support the contribution of the structural pathway and show that useful inputs are not restricted to the earliest encoder layers. Frequency interventions further identify how those inputs affect depth errors near boundaries. SPACE-CLIP is a monocular depth method designed for shared visual perception, with robotic and VLA integration motivating its architecture.

Our contributions are summarized as follows:
\begin{itemize}
\item We present SPACE-CLIP, a decoder-only depth framework that combines FiLM-conditioned semantic processing and structural refinement under the TFI-FB constraint.
\item We evaluate indoor and outdoor depth reconstruction, backbone transfer, and shared-encoder integration cost, and validate the structural pathway with a controlled component comparison.
\item We compare learned pathway representations and assess how layer selection and frequency filtering affect depth reconstruction.
\end{itemize}

\section{Related Work}
\label{sec:related_work}

\subsection{Monocular Depth Estimation}
Monocular depth estimation has progressed from CNN encoder-decoder models to Transformer- and SSM-based designs. Early CNN methods established strong supervised baselines but often lost boundary detail due to repeated downsampling \cite{9417196}. Later work improved feature fusion and objectives, including ordinal regression in DORN \cite{fu2018deep} and improved losses \cite{Bhat_2021_CVPR}.

Transformer-based methods improved long-range reasoning. DPT \cite{ranftl2021vision} showed that ViT backbones can outperform strong CNN baselines. Follow-up works such as DepthFormer \cite{li2023depthformer}, ASTransformer \cite{chang2021transformer}, and NeWCRFs \cite{yuan2022neural} further improved performance through specialized architectural choices. MambaDepth \cite{grigore2024mambadepth} explored state-space modeling for efficient long-range dependency capture.

Depth foundation models broaden the scope from dataset-specific training to cross-domain generalization. Depth Anything and Depth Anything V2 learn from large-scale pseudo-labeled data \cite{yang2024depthanything,yang2024depthanythingv2}; UniDepth and Metric3Dv2 address metric prediction across domains and camera settings \cite{Piccinelli_2024_CVPR,hu2024metric3dv2}. These advances improve the utility of depth in varied environments. The integration question addressed here is complementary: when a host system already computes visual features, a depth decoder can reuse them instead of adding another encoder.

Work on hybrid encoders, uncertainty, multimodal cues, and self-supervised objectives explores additional ways to improve depth prediction \cite{neucom_liu2025_hybriddepth,neucom_zhu2024_tsudepth,neucom_huang2024_rcdformer,neucom_zhang2020_unsupdepth,neucom_chen2020_selfsupdepth}. Lightweight monocular models and efficient attention also examine the accuracy--complexity trade-off \cite{sr_liu2025simmde,sr_schiavella2025efficient,sr_sui2024lightweight}. SPACE-CLIP focuses on the cost of adding depth to a compatible existing encoder, as well as the decoder's prediction quality.

\subsection{CLIP-Based Depth Estimation}
Large-scale VLMs such as CLIP opened new directions for depth estimation. Prior work mainly follows two routes: prompt engineering for depth querying and architectural adaptation of vision encoders.

\subsubsection{Advancements in Prompt Engineering for Depth}
Prompt-centered methods ask how to query CLIP for depth. DepthCLIP \cite{zhang2022can} used fixed prompts such as ``close'' and ``far,'' but discrete language is a weak representation of continuous geometry. Later work introduced learnable or adaptive prompts. Auty et al. \cite{Auty_2023_ICCV} replaced human-defined words with continuous tokens; CaBins \cite{son2024cabins} generated prompts from image features; Hu et al. \cite{hu2024learning} used scene-dependent codebooks; and CLIP2Depth \cite{kim2024clip} used mirror embeddings. CLIP2Depth conditions a FiLM-based decoder on text-encoded mirror features while keeping CLIP frozen. Its conditioning mechanism differs from the image--text similarity matching used by depth-bin methods.

\subsubsection{Architectural Adaptation of Vision Encoders}
Another line of work adapts CLIP vision encoders for dense prediction. For example, CaBins \cite{son2024cabins} modifies the image pathway for richer multi-scale features. However, prediction still depends on prompt-mediated image-text alignment.

SPACE-CLIP conditions FiLM directly on the pooled image representation and combines this semantic pathway with separately refined structural features. The decoder maps intermediate visual features to depth without text-derived conditioning. Component and layer-selection experiments assess the structural pathway's contribution.

\begin{figure*}[tp]
\centering
\includegraphics[width=0.96\textwidth,height=0.34\textheight,keepaspectratio]{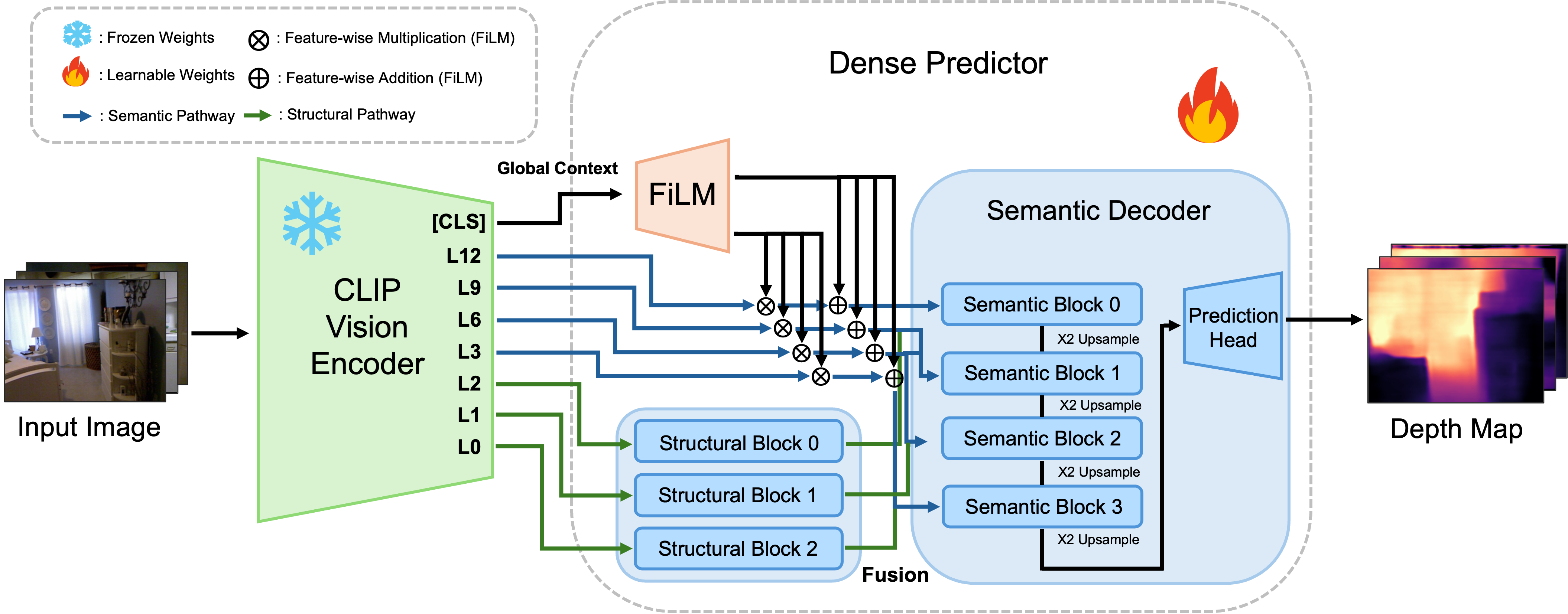}
\caption{SPACE-CLIP depth decoder. Frozen CLIP features enter FiLM-conditioned semantic processing and a structural pathway, followed by staged fusion and depth prediction.}
\label{fig:architecture}
\end{figure*}

\subsection{Intermediate Representations for Dense Prediction}
Intermediate representations give dense prediction decoders access to features from multiple encoder stages. CNN visualizations show how learned features vary across network depth \cite{zeiler2013visualizing}. In depth estimation, parallel decoding and multiscale fusion combine features for dense reconstruction \cite{sr_liu2022parallel,sr_xia2024pyramid}. These approaches motivate examining how a decoder uses features from different stages, rather than relying only on a final image representation.

Vision transformers can aggregate global information early, and their representational hierarchy differs from that of convolutional networks \cite{raghu2021dovision}. Layer depth alone therefore does not determine which features are most useful to a depth decoder. SPACE-CLIP combines FiLM-conditioned semantic processing with a separately refined structural pathway. Its layer-selection experiments examine which intermediate CLIP states support the structural pathway within this fixed decoder design.

Spatial-frequency analysis provides a complementary view of these representations. Spectral studies characterize spatial variation in transformer features \cite{park2022how}, while frequency-domain refinement has also been explored for sparse-to-dense depth completion \cite{sr_wang2026frequency}. We use frequency filtering to test how structural-pathway inputs contribute to depth prediction.

\section{Method}
\label{sec:method}
SPACE-CLIP predicts dense depth from a frozen CLIP vision encoder. As shown in Fig.~\ref{fig:architecture}, the encoder provides patch features from several layers and a global image representation. The semantic pathway uses this global context to modulate its patch features, while the structural pathway processes a separate set of encoder states. Fusion blocks combine both streams as the decoder increases spatial resolution, and a prediction head produces the depth map.

The CLIP input is generated by bicubic resizing to $224\times224$. Depth is predicted at the target image resolution, with all task-specific parameters contained in the decoder.

\subsection{Overall Architecture}
Our design principle is modularity. By freezing the large CLIP backbone and training only the Dense Predictor, SPACE-CLIP can serve as a depth module in systems that use a compatible vision encoder. Reuse requires matching encoder weights, preprocessing, and access to intermediate features.

We use pre-trained CLIP ViT-B/16 as the frozen backbone. The Dense Predictor has a Semantic pathway for context-conditioned decoding and a Structural pathway for spatial refinement. These names specify the roles of the processing branches; their inputs need not contain mutually exclusive kinds of information. Separate feature projections precede hierarchical fusion.

\subsection{Dual Pathway Feature Processing}
The Dense Predictor assigns different CLIP hidden states to the two pathways. L0 denotes the embedding output and L1--L12 the transformer-block outputs. Each provides 196 patch tokens on the same $14\times14$ grid.

\subsubsection{Semantic Pathway with FiLM}
The semantic pathway uses CLIP states L12, L9, L6, and L3. To improve context-aware decoding, we use Feature-wise Linear Modulation (FiLM) \cite{perez2018film}. Global context is extracted from the final pooled \texttt{[CLS]} representation. A small MLP maps this vector $v_{cls}$ to channel-wise scale and shift parameters $\gamma$ and $\beta$, which modulate each semantic patch feature $F_{patch}$:
\begin{equation}
\text{FiLM}(F_{patch}, \gamma, \beta) = \gamma \cdot F_{patch} + \beta
\label{eq:film}
\end{equation}
This conditioning adapts local feature interpretation to global scene context.

\subsubsection{Structural Pathway}
The structural pathway uses CLIP states L2, L1, and L0 in the default configuration. Layer selection is configurable within the same dual-pathway architecture. These features enter without FiLM, providing a separately processed source of spatial information. Structural blocks progressively upsample and refine them before fusion with the semantic stream. Section~\ref{sec:controlled_ablation} also evaluates middle and late inputs to determine how this branch depends on encoder depth.

\subsection{Hierarchical Fusion Decoder}
The decoder combines the pathways at successive resolutions so that intermediate encoder features remain available throughout reconstruction. At each stage, it upsamples the current semantic stream and concatenates it with gated semantic skip features and, at the first three stages, gated structural features. Learned convolutions combine these inputs before the next upsampling step.

Decoder resolutions increase through $28\times28$, $56\times56$, $112\times112$, and $224\times224$. Repeated fusion lets the decoder refine its prediction using encoder features at each stage, rather than reconstructing the entire map from only the deepest representation. A final prediction head outputs depth, which is resized to the target image dimensions.

\subsection{Loss Function}
Loss design is important in depth estimation because it controls how spatial relations are learned from supervision \cite{9598847}. We train SPACE-CLIP with a composite objective that balances scale-invariant accuracy and local structural consistency. The loss at the final depth output is:

\begin{equation}
\mathcal{L}_{\text{total}} = (1 - \lambda_{\text{ssim}}) \mathcal{L}_{\text{SILog}} + \lambda_{\text{ssim}} \mathcal{L}_{\text{SSIM}}
\label{eq:total_loss}
\end{equation}
where $\lambda_{\text{ssim}} = 0.5$.

\subsubsection{Scale-Invariant Logarithmic (SILog) Loss}
SILog penalizes logarithmic depth discrepancies, balancing relative structure and global scale error. For predicted depth $\hat{d}$ and ground truth depth $d$, we define $g_i = \log(\hat{d}_i) - \log(d_i)$ over $N$ valid pixels. SILog is:

\begin{equation}
\mathcal{L}_{\text{SILog}}(\hat{d}, d) = \alpha \sqrt{\frac{1}{N}\sum_{i} g_i^2 - \frac{\lambda}{N^2} \left(\sum_{i} g_i\right)^2}
\label{eq:silog}
\end{equation}
We set $\lambda=0.85$ and $\alpha=10$. With this choice, the square-root argument is $\operatorname{Var}(g)+0.15\operatorname{Mean}(g)^2$, so the loss retains sensitivity to global scale error.

\subsubsection{Structural Similarity (SSIM) Loss}
SILog captures relational depth accuracy but does not directly enforce local structural consistency. We therefore add SSIM loss, which compares luminance, contrast, and structure between predicted and ground-truth depth maps. For two windows $x$ and $y$:

\begin{equation}
\text{SSIM}(x, y) = \frac{(2\mu_x\mu_y + C_1)(2\sigma_{xy} + C_2)}{(\mu_x^2 + \mu_y^2 + C_1)(\sigma_x^2 + \sigma_y^2 + C_2)}
\label{eq:ssim}
\end{equation}
where $\mu$ and $\sigma$ denote mean and standard deviation, and $C_1, C_2$ are stabilizing constants. We compute SSIM loss as the mean of $(1-\text{SSIM})$ over valid centers. Both depth maps are normalized using the valid reference-depth range and clamped to $[0,1]$. Local moments use an $11\times11$ Gaussian window with standard deviation 1.5 before masking, with $C_1=0.01^2$ and $C_2=0.03^2$. Auxiliary SILog losses at the first two decoder outputs are added with weights 0.10 and 0.05.

\section{Experiments}
\label{sec:experiments}
We evaluate depth reconstruction on NYU Depth V2 and KITTI, decoder transfer to another frozen backbone, and shared-encoder integration cost. A controlled NYU study then tests the structural pathway and its layer inputs. Tables distinguish non-CLIP methods (N/A), text-conditioned inference with a frozen backbone (TC-FB), text-conditioned inference with backbone updates (TC-BU), and text-free inference with a frozen backbone (TFI-FB). These labels describe architecture, while training data, supervision, and evaluation settings vary across methods.

\subsection{Datasets and Evaluation}
KITTI \cite{Geiger2013IJRR} contains outdoor driving scenes with LiDAR-derived depth, and NYU Depth V2 \cite{Silberman:ECCV12} contains indoor RGB-D scenes. Input sizes are $352\times704$ for KITTI and $480\times640$ for NYU, with a $224\times224$ CLIP branch. Training uses horizontal flips and rotations up to one degree. We report AbsRel, SqRel, RMSE, logarithmic errors, and threshold accuracy $\delta_k$, the fraction of valid pixels satisfying $\max(\hat d/d,d/\hat d)<1.25^k$.

The released KITTI Eigen lists contain 23,158 training entries and 697 test entries; 652 test entries specify ground-truth depth paths. The retained aggregate metrics do not record the final evaluated sample count. The KITTI release configuration specifies Eigen cropping and a depth range of 0.001--80 m.

The benchmark tables retain the original evaluation records. Those runs used the official evaluation lists for checkpoint selection. Their exact evaluator revision was not preserved, leaving the scaling behavior of the main KITTI and NYU scores unresolved; the separate CLIP/SigLIP transfer scores are traceable to median-scaled evaluation. The records therefore have a different selection and evaluation basis from the independently validation-selected NYU study below. Cross-paper tables provide context because the cited methods also use different protocols and supervision.

For the controlled study, 24,231 supplied NYU training frames are split by scene group, merging trailing a/b capture identifiers: 21,974 images from 224 groups for training and 2,257 images from 25 groups for validation (split seed 20260926). All 654 official test images are reserved for evaluation. Test metrics use native reference-depth resolution, the NYU Eigen crop (rows [45,471), columns [41,601)), valid depths in $(0.001,10)$ m, horizontal-flip averaging, and no median scaling. Predictions are clipped to [0.001,10] m; metrics are computed per image and averaged equally.

\begin{table*}[tp]
\caption{NYU Depth V2 benchmark results. Other rows follow CaBins \cite{son2024cabins}; $\dagger$ denotes reimplementation there. Selection and crop settings are described in Section~\ref{sec:experiments}.}
\label{tab:nyu_comparison}
\centering\footnotesize
\begingroup
\renewcommand{\arraystretch}{1.15}
\setlength{\tabcolsep}{3pt}
\begin{adjustbox}{max width=\textwidth}
\begin{tabular}{l|l|c|cccccc}
\hline
\hline
\textbf{Method} & \textbf{Approach} & \textbf{Constraint} & \textbf{AbsRel$\downarrow$} & \textbf{RMSE$\downarrow$} & \textbf{log 10$\downarrow$} & \textbf{$\delta < 1.25$$\uparrow$} & \textbf{$\delta < 1.25^2$$\uparrow$} & \textbf{$\delta < 1.25^3$$\uparrow$} \\
\hline
Make3D (reported in \cite{son2024cabins}) & Unimodal & N/A & 0.349 & 1.214 & - & 0.447 & 0.745 & 0.897 \\
DORN \cite{fu2018deep} & Unimodal & N/A & 0.115 & 0.509 & 0.051 & 0.828 & 0.965 & 0.992 \\
ASTransformer \cite{chang2021transformer} & Unimodal & N/A & 0.103 & 0.374 & 0.044 & 0.902 & 0.985 & 0.997 \\
DepthFormer \cite{li2023depthformer} & Unimodal & N/A & 0.096 & 0.339 & 0.041 & 0.921 & 0.989 & 0.998 \\
NeWCRFs \cite{yuan2022neural} & Unimodal & N/A & 0.095 & 0.334 & 0.041 & 0.922 & 0.992 & 0.998 \\
\hline
DepthCLIP \cite{zhang2022can} & CLIP-based & TC-FB & 0.388 & 1.167 & 0.156 & 0.394 & 0.683 & 0.851 \\
Hu \textit{et al.} \cite{hu2024learning} & CLIP-based & TC-FB & 0.347 & 1.049 & 0.140 & 0.428 & 0.732 & 0.898 \\
Auty$^{\dagger}$ \cite{Auty_2023_ICCV} & CLIP-based & TC-FB & 0.324 & 0.961 & 0.127 & 0.473 & 0.779 & 0.921 \\
CaBins \cite{son2024cabins} & CLIP-based & TC-BU & 0.120 & 0.401 & 0.050 & 0.866 & 0.978 & 0.996 \\
\rowcolor{Gray}
SPACE-CLIP (Ours) & CLIP-based & TFI-FB & 0.1042 & 0.3848 & 0.0446 & 0.8958 & 0.9839 & 0.9973 \\
\hline
\hline
\end{tabular}\end{adjustbox}
\endgroup
\end{table*}
\begin{table*}[tp]
\caption{KITTI benchmark results (Eigen split, 0--80 m). Source methods use different supervision and evaluation protocols; SPACE-CLIP follows the original records described in Section~\ref{sec:experiments}.}
\label{tab:sota_comparison}
\centering\footnotesize
\definecolor{Gray}{gray}{0.9}
\begingroup
\renewcommand{\arraystretch}{1.15}
\setlength{\tabcolsep}{3pt}
\begin{adjustbox}{max width=\textwidth}
\begin{tabular}{l|l|c|cccc|ccc}
\hline
\hline
\textbf{Method} & \textbf{Approach} & \textbf{Constraint} & \textbf{AbsRel$\downarrow$} & \textbf{SqRel$\downarrow$} & \textbf{RMSE$\downarrow$} & \textbf{RMSE log$\downarrow$} & \textbf{$\delta < 1.25$$\uparrow$} & \textbf{$\delta < 1.25^2$$\uparrow$} & \textbf{$\delta < 1.25^3$$\uparrow$} \\
\hline
DORN \cite{fu2018deep} & Unimodal & N/A & 0.072 & 0.307 & 2.727 & 0.120 & 0.932 & 0.984 & 0.994 \\
ASTransformer \cite{chang2021transformer} & Unimodal & N/A & 0.058 & - & 2.685 & 0.089 & 0.963 & 0.995 & 0.999 \\
DepthFormer \cite{li2023depthformer} & Unimodal & N/A & 0.052 & 0.158 & 2.143 & 0.079 & 0.975 & 0.997 & 0.999 \\
NeWCRFs \cite{yuan2022neural} & Unimodal & N/A & 0.052 & 0.155 & 2.129 & 0.079 & 0.974 & 0.997 & 0.999 \\
\hline
DepthCLIP \cite{zhang2022can} & CLIP-based & TC-FB & 0.473 & 6.007 & 12.958 & - & 0.281 & 0.531 & 0.696 \\
Hu \textit{et al.} \cite{hu2024learning} & CLIP-based & TC-FB & 0.384 & 4.661 & 12.290 & - & 0.312 & 0.569 & 0.739 \\
Auty \textit{et al.} \cite{Auty_2023_ICCV} & CLIP-based & TC-FB & 0.307 & 2.197 & 6.405 & 0.121 & 0.548 & 0.826 & 0.935 \\
CLIP2Depth \cite{kim2024clip} & CLIP-based & TC-FB & 0.074 & 0.303 & 2.948 & - & 0.938 & 0.990 & 0.998 \\
CaBins \cite{son2024cabins} & CLIP-based & TC-BU & 0.057 & 0.186 & 2.322 & 0.088 & 0.964 & 0.995 & 0.999 \\
\rowcolor{Gray}
SPACE-CLIP (Ours) & CLIP-based & TFI-FB & 0.0901 & 0.4701 & 3.8451 & 0.1528 & 0.9088 & 0.9812 & 0.9945 \\
\hline
\hline
\end{tabular}\end{adjustbox}
\endgroup
\end{table*}

\subsection{Implementation Details}
\subsubsection{Model and Training}
The controlled NYU experiments use a frozen CLIP ViT-B/16 encoder \cite{dosovitskiy2020image}, semantic indices [12,9,6,3], structural indices [2,1,0], decoder channels [256,128,64,32], and residual-block dropout 0.1. AdamW uses learning rate $10^{-4}$ and weight decay 0.01 for 20 epochs, with a 5\% linear warmup and cosine decay to $10^{-6}$, batch size 32, EMA decay 0.996, and gradient clipping at norm 1.0 in the controlled study. The final SILog/SSIM objective and auxiliary losses are shared across controlled configurations.

The controlled runs use seeds 42--46. Each checkpoint is selected by the lowest validation AbsRel across raw and EMA weights; validation does not use flip averaging. The seed-42 layer runs were evaluated first and subsequently extended to seeds 43--46. The semantic-only component controls were specified after those layer-study results, with all five control trainings completed before any control test evaluation. All runs are retained. These controlled runs used an RTX 5090, PyTorch 2.10.0, and Transformers 5.2.0 in float32; encoder hashes confirmed frozen weights. Configurations, per-image results, histories, and reproduction tools are available with the code.

The KITTI release configuration used batch size 64, seed 42, and 20 epochs, with AdamW, cosine warmup, auxiliary SILog losses, and EMA. It enabled flip averaging for final evaluation only; its checkpoint-selection and evaluator limitations are described above.

\subsection{NYU Depth V2 Evaluation}
Table~\ref{tab:nyu_comparison} compares NYU depth estimates. Other methods' values come from CaBins \cite{son2024cabins}, with $\dagger$ denoting reimplementation in that source. SPACE-CLIP reports 0.1042 AbsRel under the original selection and evaluation conditions described above. Its configuration specifies no NYU crop, whereas the source baselines generally use the standard crop. The controlled experiments in Section~\ref{sec:controlled_ablation} use a common crop and independent validation selection.

\begin{figure}[tbp]
\centering
\includegraphics[width=0.97\columnwidth,height=0.44\textheight,keepaspectratio]{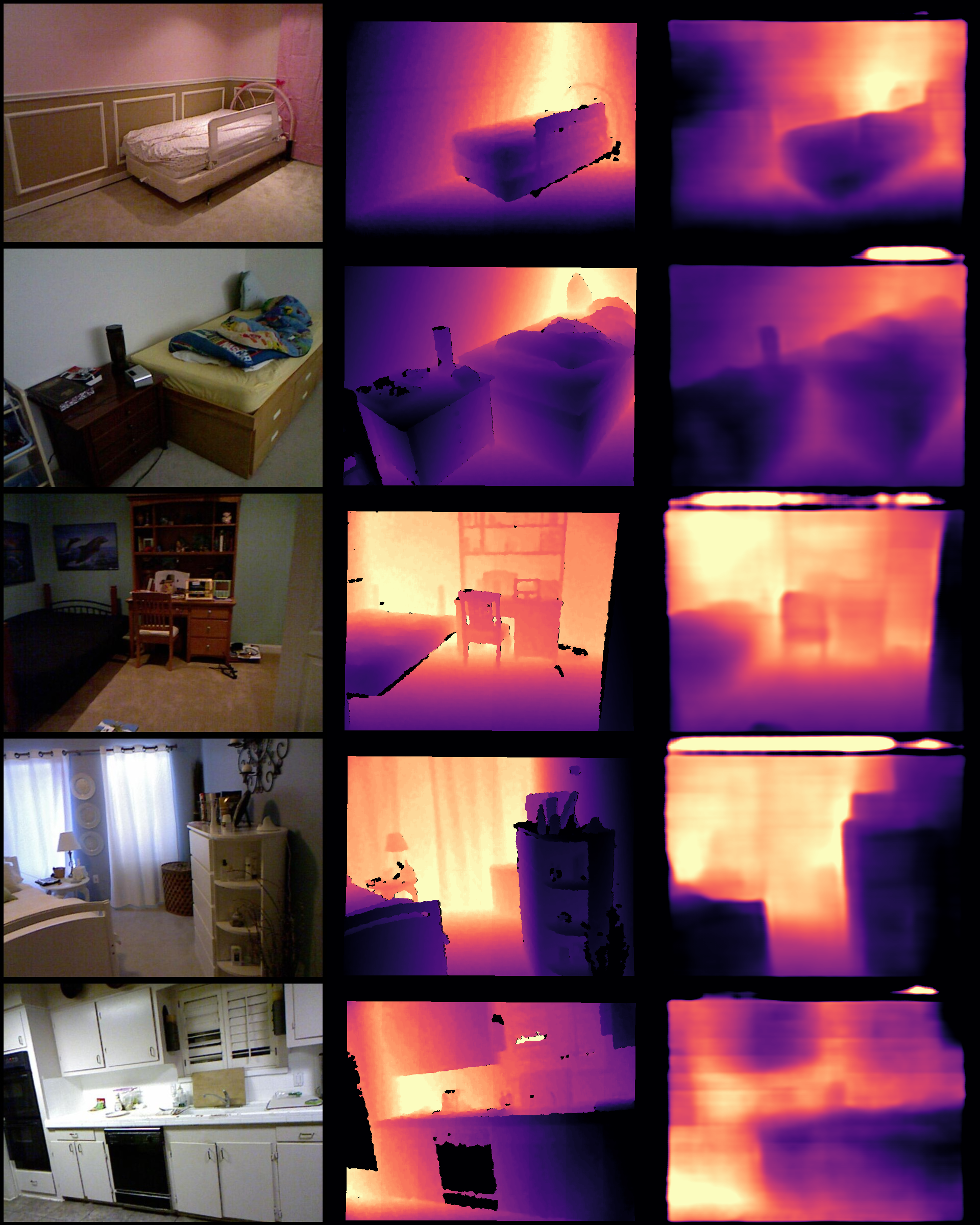}
\caption{NYU Depth V2 predictions. Columns show RGB input, reference depth, and SPACE-CLIP output.}
\label{fig:qualitative_nyu}
\end{figure}

Figure~\ref{fig:qualitative_nyu} shows indoor depth predictions that recover room layout and major object boundaries. The examples illustrate the decoder's application to dense scene reconstruction.

\subsection{KITTI Evaluation}
Table~\ref{tab:sota_comparison} reports KITTI depth estimates alongside prior monocular and CLIP-based methods. SPACE-CLIP reports 0.0901 AbsRel with the frozen-backbone, text-free design. The table places its outdoor predictions in the context of methods with different adaptation and inference requirements; the source results are not measurements from a common retraining protocol.

\begin{figure}[tbp]
\centering
\includegraphics[width=0.97\columnwidth,height=0.38\textheight,keepaspectratio]{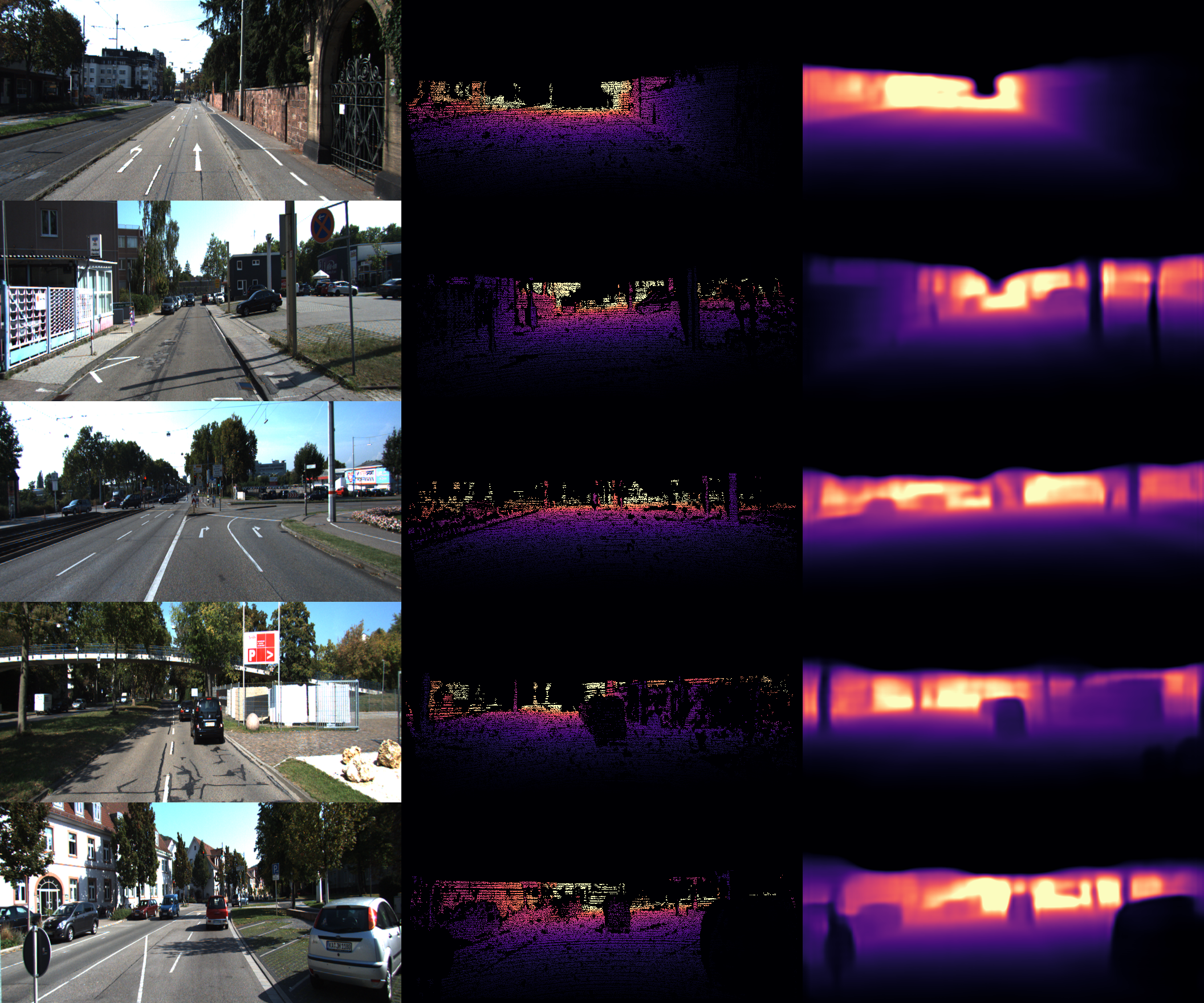}
\caption{KITTI predictions. Columns show RGB input, reference depth, and SPACE-CLIP output.}
\label{fig:qualitative_comparison}
\end{figure}

The KITTI examples in Fig.~\ref{fig:qualitative_comparison} show road layout, vehicles, and surrounding scene structure. Together with the NYU examples, they illustrate the same decoder design applied to indoor and outdoor imagery.

\subsection{Backbone Transfer on NYU}
To test whether the decoder is tied to CLIP alone, we replace frozen CLIP ViT-B/16 with frozen SigLIP ViT-B/16 \cite{zhai2023sigmoid} while retaining the dual-pathway design. Table~\ref{tab:backbone_transfer_nyu} reports median-scaled AbsRel values of 0.1037 and 0.1022. These runs demonstrate training the decoder with both backbone families. They used test-selected checkpoints; the controlled component and layer experiments use CLIP and unscaled depth evaluation.

\begin{table}[t]
\centering\footnotesize
\caption{Frozen-backbone transfer on NYU. These recorded results use median scaling and test-selected checkpoints.}
\label{tab:backbone_transfer_nyu}
\setlength{\tabcolsep}{3pt}\renewcommand{\arraystretch}{1.18}
\begin{adjustbox}{max width=\columnwidth}
\begin{tabular}{@{}lrr@{}}
\toprule
Metric & CLIP & SigLIP \\
\midrule
AbsRel & 0.1037 & 0.1022 \\
SqRel (m) & 0.0556 & 0.0556 \\
RMSE (m) & 0.3815 & 0.3776 \\
Log RMSE & 0.1366 & 0.1360 \\
$\log_{10}$ & 0.0444 & 0.0437 \\
$\delta_1$ & 0.8960 & 0.8980 \\
$\delta_2$ & 0.9831 & 0.9820 \\
$\delta_3$ & 0.9973 & 0.9963 \\
\bottomrule
\end{tabular}\end{adjustbox}
\end{table}

\begin{table}[tbp]
\centering\footnotesize
\caption{Integration cost at batch size one: backbone alone, shared encoder with SPACE-CLIP, and a separate depth encoder. All depth configurations retain TFI-FB.}
\label{tab:integration_cost}
\setlength{\tabcolsep}{3pt}\renewcommand{\arraystretch}{1.18}
\begin{adjustbox}{max width=\columnwidth}
\begin{tabular}{@{}lrrr@{}}
\toprule
Metric & Backbone & Shared & Separate \\
\midrule
Total params & 85,799,424 & 97,850,420 & 183,649,844 \\
Trainable params & 0 & 12,050,996 & 12,050,996 \\
Duplicated params & 0 & 0 & 85,799,424 \\
Peak memory (MB) & 346.9 & 468.1 & 796.1 \\
Latency (ms/image) & 3.368 & 5.200 & 8.512 \\
Shared encoder & Yes & Yes & No \\
\bottomrule
\end{tabular}\end{adjustbox}
\end{table}

\subsection{System-Level Integration Cost}
We compare a frozen backbone alone, a shared backbone with SPACE-CLIP, and a duplicated-backbone depth stack at batch size one. Each configuration runs in an isolated subprocess with 10 warm-up and 50 timed iterations. Table~\ref{tab:integration_cost} reports 5.200 ms/image and 468.1 MB for sharing, versus 8.512 ms/image and 796.1 MB for duplication. Sharing removes 85,799,424 duplicated backbone parameters. GPU-model metadata was not retained for this separate microbenchmark. These measurements describe its inference setup; end-to-end robotic control is outside this benchmark.

\begin{table}[tbp]
\centering\footnotesize
\caption{KITTI component configurations. Full uses a different training recipe from Base, +FiLM, and +Struct.; Table~\ref{tab:controlled} uses a shared NYU recipe.}
\label{tab:ablation_study}
\setlength{\tabcolsep}{3pt}\renewcommand{\arraystretch}{1.18}
\begin{adjustbox}{max width=\columnwidth}
\begin{tabular}{@{}lrrrr@{}}
\toprule
Metric & Base & +FiLM & +Struct. & Full \\
\midrule
FiLM & -- & $\checkmark$ & -- & $\checkmark$ \\
Structural path & -- & -- & $\checkmark$ & $\checkmark$ \\
AbsRel & 0.1165 & 0.1142 & 0.1094 & 0.0901 \\
SqRel (m) & 0.7981 & 0.7684 & 0.7238 & 0.4701 \\
RMSE (m) & 5.152 & 5.121 & 5.143 & 3.8451 \\
Log RMSE & 0.1962 & 0.1941 & 0.1911 & 0.1528 \\
$\delta_1$ & 0.858 & 0.862 & 0.870 & 0.9088 \\
$\delta_2$ & 0.963 & 0.964 & 0.966 & 0.9812 \\
$\delta_3$ & 0.988 & 0.988 & 0.989 & 0.9945 \\
\bottomrule
\end{tabular}\end{adjustbox}
\end{table}

\begin{table*}[tp]
\centering\footnotesize
\caption{Controlled NYU test results: mean $\pm$ sample SD over seeds 42--46, with all 654 images, validation-selected checkpoints, and no median scaling. Boundary AbsRel uses threshold 0.05. All configurations retain FiLM.}
\label{tab:controlled}
\setlength{\tabcolsep}{4pt}\renewcommand{\arraystretch}{1.2}
\begin{adjustbox}{max width=\textwidth}
\begin{tabular}{@{}llrrrrr@{}}
\toprule
Model & Structural layers & \shortstack{Trainable\\params (M)} & AbsRel $\downarrow$ & RMSE (m) $\downarrow$ & $\delta_1$ $\uparrow$ & \shortstack{Boundary\\AbsRel $\downarrow$} \\
\midrule
Semantic only & -- & 7.57 & 0.1412 $\pm$ 0.0184 & 0.5121 $\pm$ 0.0738 & 0.7797 $\pm$ 0.0745 & 0.1583 $\pm$ 0.0169 \\
SPACE-CLIP (early) & 2, 1, 0 & 12.05 & 0.1283 $\pm$ 0.0113 & 0.4544 $\pm$ 0.0462 & 0.8325 $\pm$ 0.0389 & 0.1447 $\pm$ 0.0103 \\
SPACE-CLIP (middle) & 7, 5, 4 & 12.05 & 0.1208 $\pm$ 0.0017 & 0.4355 $\pm$ 0.0080 & 0.8533 $\pm$ 0.0073 & 0.1393 $\pm$ 0.0015 \\
SPACE-CLIP (late) & 11, 10, 8 & 12.05 & 0.1230 $\pm$ 0.0020 & 0.4418 $\pm$ 0.0054 & 0.8487 $\pm$ 0.0057 & 0.1432 $\pm$ 0.0023 \\
\bottomrule
\end{tabular}\end{adjustbox}
\end{table*}

\subsection{Ablation Studies}
\subsubsection{FiLM and Structural Components on KITTI}
Table~\ref{tab:ablation_study} compares component configurations on KITTI. The semantic-only baseline without FiLM gives AbsRel 0.1165; adding FiLM gives 0.1142, and adding the structural pathway without FiLM gives 0.1094. The full model reports 0.0901. The full-model training recipe differs from the reduced arms, so these values motivate the combined design without isolating a FiLM--pathway interaction. We assess the structural pathway under a shared training recipe in the following NYU comparison.

\subsubsection{Controlled Structural-Pathway Comparison}
\label{sec:controlled_ablation}
The semantic-only control keeps FiLM and semantic indices [12,9,6,3], but removes the structural pathway. For each seed, shared tensors are copied from the corresponding full-model initialization; the three affected fusion convolutions retain the input-channel slices for the semantic stream and skip features without rescaling. Splits, losses, minibatch order, optimizer, and schedule are shared. Removing the pathway reduces trainable parameters from 12,050,996 to 7,570,932 and can change dropout random-number consumption. The experiment evaluates the full architecture against this smaller decoder.

Table~\ref{tab:controlled} reports mean and sample standard deviation over five seeds. The default early-layer model reduces mean test AbsRel from 0.1412 to 0.1283, a 9.2\% relative decrease, with lower error in all five seeds. Same-seed full-minus-control differences average $-0.0130\pm0.0129$. The largest gains occur in the less stable control runs at seeds 44 and 45; all five remain in the aggregate.

To examine local reconstruction, we define reference boundaries by adjacent valid-pixel log-depth jumps exceeding 0.05 and dilate them by three pixels. Eroding the validity mask by the same radius excludes missing-depth and crop-edge neighborhoods. The early model lowers boundary-neighborhood AbsRel from 0.1583 to 0.1447, again in all five seeds. The comparison supports adding the structural pathway under this recipe, while its parameter-count difference prevents attributing the gain solely to pathway organization.

\subsubsection{Structural-Pathway Layer Selection}
We compare early [2,1,0], middle [7,5,4], and late [11,10,8] inputs while keeping the semantic pathway, architecture, and parameter count fixed. Within each seed, the three models start from identical decoder weights and follow the same minibatch order. Middle gives the lowest mean whole-image AbsRel, followed by late and early (Table~\ref{tab:controlled}). One less stable early run at seed 44 contributes substantially to its larger standard deviation.

Middle has lower whole-image error than late in three of five runs, and lower boundary-neighborhood error in all five. Boundary AbsRel is 0.1393 for middle and 0.1432 for late; the same mean ordering holds at boundary thresholds 0.03 and 0.10. Middle and late also improve whole-image error over the semantic-only control in all five runs, with boundary improvements in five and four runs, respectively. These observations favor selecting structural-pathway inputs by regional accuracy and repeatability, rather than assuming that only shallow features support geometry.

\subsection{Analysis of Pathway Features}
\subsubsection{Learned Features Before and After Fusion}
\begin{figure*}[tp]
\centering
\includegraphics[width=0.80\textwidth,height=0.42\textheight,keepaspectratio]{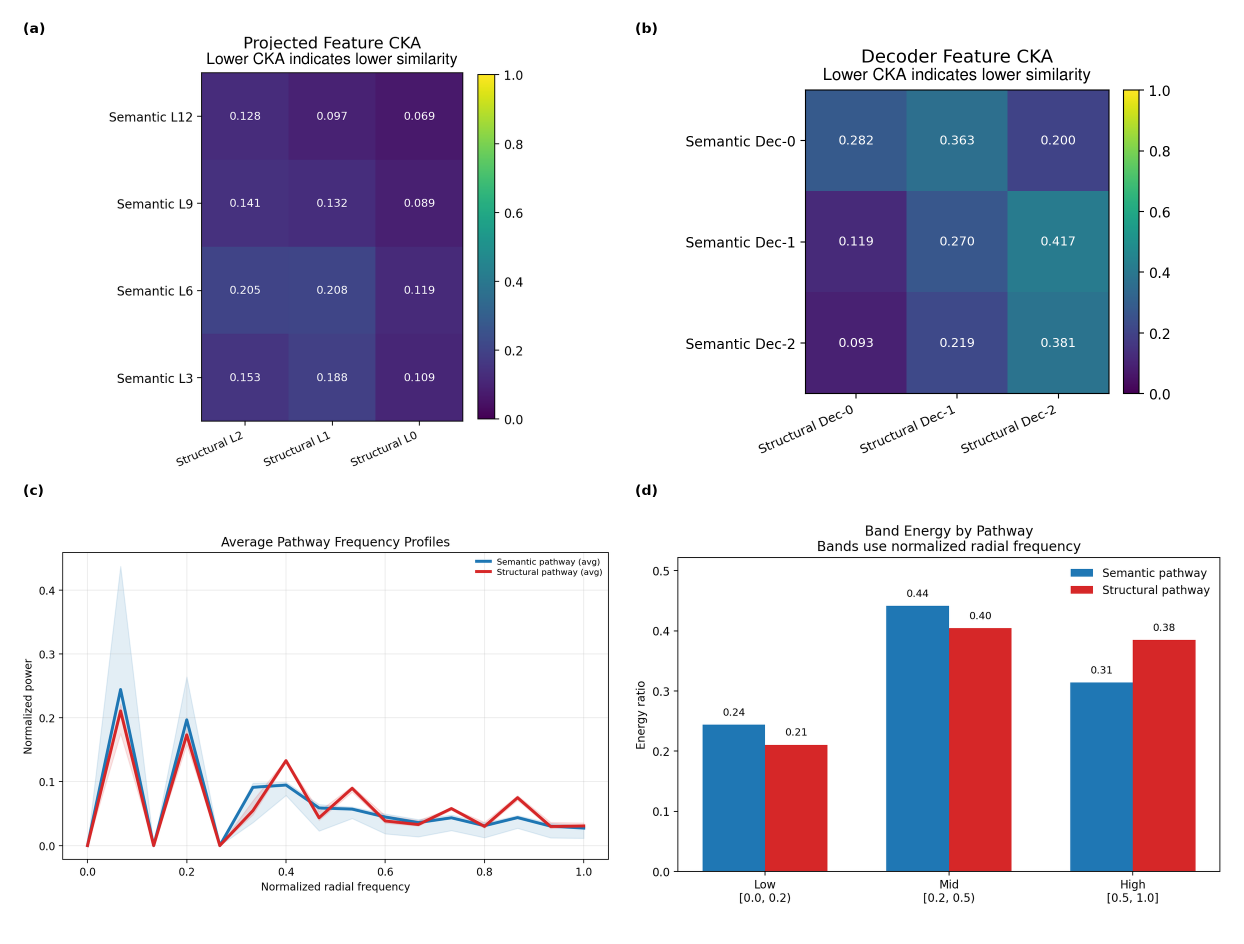}
\caption{Learned pathway features: (a,b) cross-path CKA before and after fusion; (c,d) radial spectra and band energy. Lower CKA denotes lower representation similarity.}
\label{fig:analysis_specialization}
\end{figure*}

Figure~\ref{fig:analysis_specialization} compares learned pathway features on NYU Depth V2. We use linear centered kernel alignment (CKA) \cite{kornblith2019similarity} to compare semantic projections from L12, L9, L6, and L3 with structural projections from L2, L1, and L0, averaging over held-out images. The semantic projections already include FiLM conditioning. Decoder-stage CKA compares the learned semantic and structural feature maps after fusion. Mean cross-path CKA is 0.1365 before fusion and 0.2603 after fusion, indicating greater representation similarity after the streams interact.

Radial power spectra summarize the learned pathway features using normalized radial frequency from zero to one, with band cutoffs at 0.2 and 0.5. The reported high-frequency energy fractions are 0.3847 for the structural pathway and 0.3140 for the semantic pathway. These measurements describe pathway representations without implying an exclusive division of semantic and geometric information. The normalized bands differ from the native-grid cutoff used in the filtering experiment below.

\begin{figure*}[tp]
\centering
\includegraphics[width=0.94\textwidth,height=0.31\textheight,keepaspectratio]{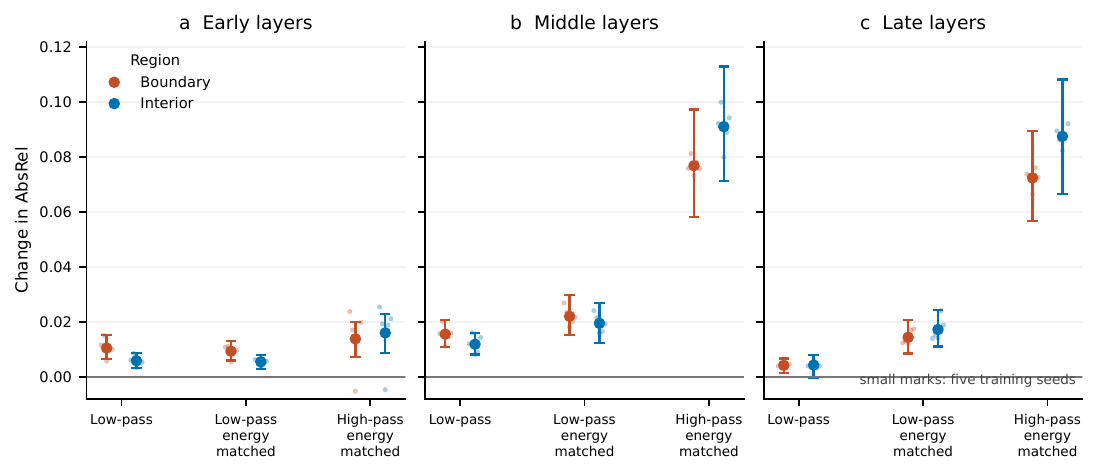}
\caption{Frequency removal from structural-pathway inputs. Large points show five-run means, small marks individual runs, and bars 95\% scene-bootstrap intervals. Positive changes indicate increased AbsRel.}
\label{fig:perturbation}
\end{figure*}

\subsubsection{Frequency Sensitivity of Depth Predictions}
We evaluate frequency sensitivity on 200 validation images: eight fixed frames from each of 25 scene groups. Spatial frequencies are measured on the $14\times14$ patch grid, with $r$ denoting radial frequency in cycles per input field. Evaluation uses no flip averaging; image-level metrics are averaged within each scene and scenes receive equal weight.

For each validation-selected dual-pathway model, we filter only the structural pathway's selected patch features, keeping semantic features, pooled context, and model weights fixed. Low-pass filtering removes $r\geq4$; high-pass filtering removes $0<r<4$. Energy-matched conditions rescale retained non-DC (AC) components to their original total energy while preserving channel means. Fourier operations use float64 before casting back to the original dtype. An FFT round-trip control gives zero maximum prediction difference across all 3,000 image--model pairs.

Figure~\ref{fig:perturbation} reports paired changes from each model's unfiltered prediction. Removing high frequencies from early inputs increases boundary AbsRel by $0.0105\pm0.0034$ across training runs, or $0.0093\pm0.0023$ after energy matching. Interior increases are 0.0058 and 0.0054, respectively; the boundary-minus-interior contrast is positive in every run. The pooled boundary effect has 95\% scene-bootstrap intervals [0.0065,0.0153] and [0.0059,0.0131]. For these intervals, paired per-image effects are first averaged over runs, then over the 25 scenes, with 2,000 scene resamples.

Sensitivity varies with layer selection. Energy-matched high-frequency removal increases mean boundary error by 0.0220 for middle and 0.0143 for late. Conversely, removing low frequencies while retaining energy-matched high frequencies increases whole-image AbsRel by 0.0154, 0.0893, and 0.0849 for early, middle, and late. The decoder thus depends on multiple frequency ranges, with different balances across layer choices. These exploratory validation-scene interventions characterize trained-model sensitivity; the intervals are unadjusted and are not confirmatory significance tests.

\section{Discussion}
\label{sec:discussion}
\subsection{Interpretation of Results}
SPACE-CLIP adapts a frozen visual encoder through a decoder that combines context-conditioned semantic processing and structural refinement. The controlled component comparison supports this architecture under the tested training recipe: the full early-layer model lowers whole-image and boundary error in every paired run. Layer selection refines this architecture: the middle-layer configuration gives the lowest mean error under the controlled NYU protocol, while early and late states also provide useful structural-pathway inputs.

The feature analyses describe the processing roles within the decoder. Cross-path similarity increases after fusion, and removing high frequencies from early structural inputs affects boundary neighborhoods more strongly than interiors. Both low- and high-frequency components contribute to prediction, with sensitivity depending on layer selection.

\subsection{Modular Perception for Embodied Systems}
In robotic perception, a depth model's usefulness also depends on how it attaches to a larger system. Monocular manipulation, visual servoing, and VLM-assisted grasping illustrate settings that connect visual understanding to action \cite{ras_haugalokken2020_monocular,ras_ribeiro2021_visualservo,ras_alshanoon2022_servo,ras_kim2026_vlmgrasp,ras_zhang2026_ovgrasp}. SPACE-CLIP addresses the representation interface by keeping the vision backbone frozen and learning depth-specific processing entirely within the decoder.

The shared-backbone benchmark quantifies the cost of avoiding a second image encoder. Reuse is applicable when the host provides the same encoder states and preprocessing; it is not automatic for a fine-tuned or differently configured VLA model. The integration in Fig.~\ref{fig:conceptual_paradigm} is a proposed use of the depth module. Its effect on policy learning, end-to-end latency, and robotic control requires downstream evaluation.

\subsection{Limitations and Future Work}
The controlled experiments cover one indoor dataset and one frozen backbone. All seeds are retained, including less stable runs. The component control has fewer parameters, and its comparison was added after observing the layer-study results; it does not isolate capacity or establish FiLM--pathway synergy. The original KITTI, NYU, and backbone-transfer tables have the evaluation limitations described in Section~\ref{sec:experiments}, so the controlled NYU study provides the independent-validation evidence.

Frequency interventions use validation scenes and can introduce distribution shifts or ringing. Scene-bootstrap intervals and across-seed standard deviations describe different sources of variation. Future work should extend the controlled protocol to outdoor data and additional encoders, examine capacity-matched decoder alternatives, and evaluate shared-encoder depth in robotic and VLA systems under adverse sensing conditions.

\section{Conclusion}
\label{sec:conclusion}
We introduced SPACE-CLIP, a decoder-only monocular depth framework that combines FiLM-conditioned semantic features and structural refinement from a frozen CLIP vision encoder. It predicts dense depth without a text encoder at inference and supports reuse of a compatible visual backbone. Indoor and outdoor evaluations illustrate the decoder's application, and controlled component comparisons support the structural pathway. Feature comparisons and frequency interventions characterize the learned representations and their use in reconstruction. The shared-backbone microbenchmark further demonstrates reduced parameter duplication and inference cost in the tested setup. These results support decoder-side adaptation as a practical route to adding spatial prediction to existing visual perception systems.
\section*{Acknowledgment}
This work was supported by the Gachon University research fund (GCU-202500670001).

\bibliographystyle{IEEEtran}
\bibliography{reference}
\end{document}